# Recursive Partitioning for Personalization using Observational Data

Nathan Kallus [1]

## Abstract

We study the problem of learning to choose from $m$ discrete treatment options (*e.g.*, news item or medical drug) the one with best causal effect for a particular instance (*e.g.*, user or patient) where the training data consists of passive observations of covariates, treatment, and the outcome of the treatment. The standard approach to this problem is regress and compare: split the training data by treatment, fit a regression model in each split, and, for a new instance, predict all $m$ outcomes and pick the best. By reformulating the problem as a single learning task rather than $m$ separate ones, we propose a new approach based on recursively partitioning the data into regimes where different treatments are optimal. We extend this approach to an optimal partitioning approach that finds a globally optimal partition, achieving a compact, interpretable, and impactful personalization model. We develop new tools for validating and evaluating personalization models on observational data and use these to demonstrate the power of our novel approaches in a personalized medicine and a job training application.

## 1. Introduction

Personalization is the problem of determining the best treatment option for a given instance. A treatment can, for example, be a movie recommendation (Zhou et al., 2008), a display ad (Goldfarb & Tucker, 2011), or a pharmacological therapy (Lesko, 2007), and an instance is usually an individual person. In this paper, we study the problem of learning how to personalize from *observational* data, which is an important problem in emergent contexts such as personalized medicine. In this and related contexts, experimentation can be prohibitively small-scale, costly, dangerous, and unethical in comparison to passive data collection, which can be potentially massive as in electronic medical records (EMRs) but, at the same time, lack experimental manipulation so that isolated *causal* effects of specific treatments are hidden by confounding factors. We show that standard approaches that pose the problem as multiple supervised learning tasks fall short in this setting and propose new learning algorithms as well as evaluation methods used for validation, selection, and tuning.

[1]School of Operations Research and Information Engineering and Cornell Tech, Cornell University. Correspondence to: Nathan Kallus <kallus@cornell.edu>.

Specifically, we consider the problem of learning how to assign the best of $m$ treatments to an instance, given an observation of associated baseline covariates $x \in \mathbb{R}^d$. An instance is characterized by the random variables $X \in \mathbb{R}^d$ and $Y(1), \ldots, Y(m) \in \mathbb{R}$, which denote the covariates and the $m$ *potential* outcomes of applying each of the treatments (Imbens & Rubin, 2015, Chs. 1-2). We use the convention that *smaller outcome is better*. A personalization model is a map $\tau : \mathbb{R}^d \to [m] = \{1, \ldots, m\}$ that, given an observation of covariates $x$, prescribes a treatment $\tau(x)$. Its (out-of-sample) personalization risk is its average causal effect in the population $R(\tau) = \mathbb{E}\left[Y(\tau(X))\right]$ (the expectation is taken with respect to the joint distribution of $X, Y(1), \ldots, Y(m)$). The Bayes optimal risk is $R^* = R(\tau^*)$, where $\tau^*(x) \in \mathcal{T}^*(x) = \arg\min_{t \in [m]} \mathbb{E}\left[Y(t) \mid X = x\right]$ is the Bayes optimal personalization model.

The learning task is to train a personalization model $\hat{\tau}_n(\cdot)$ on $n$ data points: $S_n = \{(X_1, T_1, Y_1), \ldots, (X_n, T_n, Y_n)\}$, where the observed outcome $Y_i = Y_i(T_i)$ corresponds *only* to the treatment $T_i$ administered. This data is *observational*: we may not control the historic administration of treatment (as we would in a controlled experiment) and the values $Y_i(t)$ for $t \neq T_i$ are missing data. We assume the data is independent and identically distributed (iid) and let $X, T, Y, Y(1), \ldots, Y(m)$ represent a generic draw. Although the data is iid, the $t$-treated sample $\{i : T_i = t\}$ differs *systematically* from the sample $t'$-treated $\{i : T_i = t'\}$ for $t \neq t'$, *i.e.*, not just by chance as in a randomized controlled trial (RCT). Our second assumption about the data is unconfoundedness:

**Assumption 1.** For each $t \in [m]$: $Y(t)$ is independent of $T$ given $X$ and $T = t$ is possible for almost every $X$, *i.e.*, $Y(t) \perp\!\!\!\perp T \mid X$ and $\mathbb{P}\left(\mathbb{P}\left(T = t \mid X\right) > 0\right) = 1$.

The assumption is standard in causal effect estimation

(see *e.g.* Kallus, 2016; 2017b) for ensuring identifiability (Rosenbaum & Rubin, 1983). It says that we measured the right covariates to separate the effect of the treatment itself from the effect of assignment. In the context of personalized medicine, this assumption would be justified if the EMR contained all the patient information used by a doctor to prescribe treatment up to the vagaries and idiosyncrasies of individual doctors or hospitals. Under Asn. 1, the conditional causal effect is equal to regressing $Y$ on $X, T$:

$$\mathbb{E}\left[Y(t) \mid X = x\right] = \mathbb{E}\left[Y(t) \mid X = x, T = t\right]$$
$$= \mathbb{E}\left[Y \mid X = x, T = t\right].$$

### 1.1. Standard Approach: Regress and Compare

Since under Asn. 1 the optimal model $\tau^*$ chooses a treatment by minimizing among $m$ regression functions, one obvious approach to personalization is to estimate these regression functions, fitting each to the subset of the data that received each treatment, and then use these to predict outcomes and pick the smallest prediction. For example, in medicine, there is a vast literature on predicting patient-specific responses to treatment (Feldstein et al., 1978; Stoehlmacher et al., 2004) and picking the best by comparing (Qian & Murphy, 2011; Bertsimas et al., 2017).

The same approach is also generally taken in the contextual multi-arm bandit problem (Li et al., 2010), which is similar to our problem with the differences that we consider an offline learning problem and that bandit arm pulls are controlled interventions (a bandit problem is essentially a dynamic RCT) so the treated subpopulations are always statistically equivalent. The standard solution is to fit $m$ regression functions, and, for a new instance, predict $m$ outcomes and pick the smallest prediction subject to cleverly ensuring sufficient exploration by, *e.g.*, adding confidence bounds that vanish with $n$. The regression, assumed linear, is done using ridge regression as in Li et al. (2010) (LinUCB), ordinary least squares (OLS) as in Goldenshluger & Zeevi (2013), or LASSO as in Bastani & Bayati (2016).

The regress and compare (R&C) approach to personalization from observational data can be summarized as:

1. For each $t \in [m]$:
   (a) Consider the $t$-treated subsample $S_{t,n_t} = \{(X_i, Y_i) : i \in [n], T_i = t\}$ of size $n_t = \sum_{i=1}^{n} \mathbb{I}[T_i = t]$.
   (b) Fit a regression model $\hat{\mu}_{t,n_t}(x)$ of the response $Y$ to regressors $X$ using training data $S_{t,n_t}$, *e.g.*, by OLS.[1]

---

[1]Note that separate OLS on each subsample $S_{t,n_t}$ is equivalent to OLS on the whole sample if we include interaction terms with dummy variables $\mathbb{I}[T_i = t]$. At the same time, OLS on the whole sample without interaction terms provides no personalization, *i.e.*, $\hat{\tau}_n^{R\&C}(x)$ is constant. Similarly, separate nonparametric regressions on each subsample is equivalent to using the whole sample and endowing the $T$ variable with a discrete topology.

2. Personalize by choosing the best predicted outcome: $\hat{\tau}_n^{R\&C}(x) = \arg\min_{t \in [m]} \hat{\mu}_{t,n_t}(x)$.

Under Asn. 1, if our regression estimator is consistent then so is R&C personalization consistent as shown below. All proofs are given in the supplementary materials.

**Theorem 1.** *If Asn. 1 holds and $\hat{\mu}_{t,n_t}(x)$ are pointwise consistent regressions, i.e., $\hat{\mu}_{t,n_t}(X) \to \mathbb{E}[Y \mid X, T = t]$ almost surely (a.s.) $\forall t \in [m]$, then $\hat{\tau}_n^{R\&C}(X) \in \mathcal{T}^*(X)$ eventually a.s.*

Examples of pointwise consistent regression estimators are $k$-nearest neighbors ($k$NN) and kernel regression (Walk, 2010). If a linear model is well-specified, then OLS is also pointwise consistent. In practice, however, R&C is not effective for personalization because it attempts to learn much more than it needs to, it splits the training data into $m$, and in training it addresses estimation or prediction risk rather than personalization risk.

### 1.2. Other Related Problems and Approaches

In learning heterogeneous causal effects, one is concerned with the case of observational data with two treatments, $t = 1$ ("Control") and $t = 2$ ("Treatment"), and the estimation of the relative conditional average treatment effect (CATE), $\delta(x) = \mathbb{E}[Y(2) - Y(1) \mid X = x]$. Under Asn. 1, CATE is the difference between two regression functions: $\delta(x) = \mathbb{E}[Y \mid X = x, T = 2] - \mathbb{E}[Y \mid X = x, T = 1]$. And so one way to estimate it is by regressing outcome in each treatment population and taking differences. When the conditioning variables in CATE are a proper subset of the covariates needed to satisfy Asn. 1, Abrevaya et al. (2015) propose estimates based on propensity-score weighting and kernel regression. Athey & Imbens (2016) develop the Causal Tree (CT), which uses recursive partitioning, as an alternative to differencing two CART regressions.

For personalization, learning heterogeneous effects can be used to choose between two treatments by comparing an estimate of their relative CATE to zero. As a learning problem, however, this addresses estimation risk rather than personalization risk and deals only with two treatments. In Sec. 2.6 we propose one-vs-one and one-vs-all strategies for personalization using CATE estimates and show it is consistent. We compare to this strategy using CT in our empirical investigation.

In learning from logged bandit feedback (Beygelzimer & Langford, 2009; Swaminathan & Joachims, 2015a;b; Kallus, 2017a), one is concerned with learning a good policy for a contextual multi-arm bandit problem based on logged data from another, known policy, rather than online interactions. This problem differs in that it assumes the data is experimental and the policy that generated the data is known and available. In Sec. 2.6, we discuss adapt-ing these methods using imputed estimated propensities, to which we compare in our empirical investigation.

## 2. Recursive Partitioning for Personalization

In this section we present three new algorithms that tackle personalization directly as a single learning task.

### 2.1. Recasting the Problem

We begin by reformulating personalization risk. Following Hirano & Imbens (2004, Def. 2.1), we define the generalized propensity score (GPS) as $Q = \phi(T, X)$, where $\phi(t, x) = \mathbb{P}(T = t \mid X = x)$. The GPS of subject $i$, $Q_i$, is an *unknown* quantity given by taking the unknown $\phi(t, x)$ and plugging in the known variables $T_i, X_i$. Using the GPS we can relate the personalization risk of a personalization model $\tau$ to its accuracy as a classification model for labels $T$, weighted by outcome and GPS.

**Theorem 2.** *Under Asn. 1,*

$$R(\tau) = \mathbb{E}\left[\mathbb{I}[T = \tau(X)] Y/Q\right]. \quad (1)$$

For $m = 2$ and randomized data ($\phi(1, x) = \pi$ constant), Zhao et al. (2012) is a special case of Thm. 2. Thm. 2 suggests using a weighted form of empirical classification risk minimization. When $Q$ is fully known as in the logged bandit setting, this approach is closely related to the approach taken by Beygelzimer & Langford (2009); Swaminathan & Joachims (2015a;b). In the observational setting, we explore estimating and imputing $Q$ to use this approach in Sec. 2.6. However, because estimating the GPS generally either relies heavily on model specification or, in nonparametric settings, can be biased and variable, this will lead to severe instability and limited practical use. Moreover, it does not address the personalization problem as a single learning task, rather as two: learning a propensity model task and then a weighted classification task. Instead, in the following sections we present a single-task approach that does not rely on estimating propensities separately.

### 2.2. An Impurity Measure for Personalization

Classification and regression trees (CART) are predictive models based on recursive partitioning: the covariate space is recursively partitioned by axis-aligned hyperplanes ($x_\ell \leq \theta$ for $\ell \in [d]$ and $\theta \in \mathbb{R}$) in order to minimize a within-partition impurity measure (Breiman et al., 1984). Impurities for classification include entropy and Gini and for regression include sum of squared errors. Athey & Imbens (2016) develop impurities for estimating heterogeneous effects. Motivated by Thm. 2, we develop an impurity for personalization leading to a recursive partitioning algorithm called personalization tree (PT).

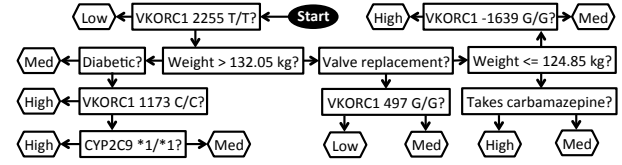

*Figure 1.* Personalization Tree for Warfarin Dosing (Sec. 4.1): from each node, the first arrow clockwise starting from noon corresponds to "No" and the other to "Yes."

Note that for a subset $\mathcal{X} \subseteq \mathbb{R}^d$ such that $X \perp\!\!\!\perp T \mid X \in \mathcal{X}$, we have that $\phi(t, x) = \mathbb{P}(T = t \mid X \in \mathcal{X})$ whenever $x \in \mathcal{X}$. To develop the personalization impurity, we use this to establish the following as a corollary of Thm. 2:

**Corollary 3.** *Consider a fixed partition of the covariate space:* $\mathbb{R}^d = \mathcal{X}_1 \cup \cdots \cup \mathcal{X}_L$ *where* $\mathcal{X}_\ell \cap \mathcal{X}_{\ell'} = \varnothing$ *whenever* $\ell \neq \ell'$. *Suppose that the partition is sufficiently fine so that*

$$X \perp\!\!\!\perp T \mid X \in \mathcal{X}_\ell \quad \forall \ell = 1, \ldots, L. \quad (2)$$

*Then,* $\hat{R}_n(\tau) = \sum_{\ell=1}^L \hat{R}_{n, \mathcal{X}_\ell}(\tau)$ *is an unbiased and consistent estimator for* $R(\tau)$, *where*

$$\hat{R}_{n, \mathcal{X}}(\tau) = \frac{\sum_{i=1}^n \mathbb{I}[X_i \in \mathcal{X}]}{n} \sum_{i=1}^n \frac{\mathbb{I}[X_i \in \mathcal{X}, T_i = \tau(X_i)] Y}{\sum_{i'=1}^n \mathbb{I}[X_{i'} \in \mathcal{X}, T_{i'} = \tau(X_i)]}.$$

Note that the conditional independence in the condition (2), which requires that leaf membership be a balancing score as defined by Rosenbaum & Rubin (1983), holds trivially when we condition on $X$ itself. Therefore, the finer the partition $\mathcal{X}_1, \ldots, \mathcal{X}_L$, the more accurate this assumption. (while possibly not truly satisfiable by any finite partition). Given a partition, we can use this risk estimate to optimize $\tau$. Letting $\tilde{S}_\ell = \{(X_i, T_i, Y_i) : X_i \in \mathcal{X}_\ell\}$, we can rewrite

$$\min_{\tau:\mathbb{R}^d \to [m]} n \sum_{\ell=1}^L \hat{R}_{n, \mathcal{X}_\ell}(\tau) = \sum_{\ell=1}^L I_{\text{pers}}(\tilde{S}_\ell),$$

where $I_{\text{pers}}$ is the personalization impurity given by

$$I_{\text{pers}}(\{(X_{i_1}, T_{i_1}, Y_{i_1}), \ldots, (X_{i_k}, T_{i_k}, Y_{i_k})\})$$
$$= \min_{t \in [m]} \frac{k(\sum_{j=1}^k \mathbb{I}[T_{i_j} = t] Y_{i_j})}{\sum_{j=1}^k \mathbb{I}[T_{i_j} = t]},$$

Therefore, to achieve good personalization risk we may wish to seek partitions that have minimal sum of within-partition personalization impurities as defined by $I_{\text{pers}}$.

### 2.3. Personalization Tree

The PT algorithm attempts to find a fine partition of the data to minimize the sum of within-partition personalization impurities. It does so by partitioning the dataset along axis-aligned cuts, at each stage choosing to cut a partition into the two partitions with least sum of impurities and proceeding recursively. In an attempt to find a fine partition that satisfies condition (2), PT continues to recurse until a specified stopping criterion. One criterion is that there be at least $n_{\text{min-leaf}} \geq 1$ subjects of each treatment in the partition.[2] Another criterion may be a maximum recursion depth $\Delta_{\max}$, but this criterion is not necessary. We also allow for a limited number $\#_{\text{features}}$ of features to be sampled as candidate cut dimensions. We summarize the PT recursive subroutine as Alg. 1. The PT algorithm is given by passing the whole dataset $S_n$ and initial depth $\Delta = 0$ to Alg. 1. The PT algorithm is notable for producing an interpretable decision tree for the personalization rule (Fig. 1).

### 2.4. Personalization Forest

The finer the partition produced by PT, the closer we are to condition (2) and the less bias the estimate of risk has. The coarser the partition, the less variance the estimate has. Therefore, there is an inherent trade-off to the fineness parameters in PT. To address this we can bag (bootstrap aggregate) many very fine PTs, which will have the effect of reducing variance without incurring too much bias as in the case of random forests (Breiman, 2001). The corresponding personalization forest (PF) algorithm is summarized in Alg. 2. Generally, we set $\#_{\text{features}}$ to $\sqrt{d}$ to achieve sufficient independence between trees for variance reduction.

### 2.5. Optimal Personalization Tree

PT is a *greedy* algorithm for minimizing personalization impurity. In this section we propose the optimal personalization tree (OPT) algorithm, which solves the *global* problem of finding partitions that minimize the sum of within-partition personalization impurities:

$$\min_{\mathcal{X}_1 \cup \cdots \cup \mathcal{X}_L = \mathbb{R}^d : (*)} \sum_{\ell=1}^{L} I_{\text{pers}}(\{(X_i, T_i, Y_i) : X_i \in \mathcal{X}_\ell\}), \quad (3)$$

where $(*)$ is the restriction that $\mathcal{X}_1, \ldots, \mathcal{X}_L$ be disjoint regions defined by the leaves of a binary decision tree. In the case of classification and regression, Bennett & Blue (1996); Bertsimas & Dunn (2016) attempt to find globally optimal prediction trees (while the problem is NP-hard; see Hyafil & Rivest, 1976). For our personalization problem, motivated by Bertsimas & Dunn (2016), we propose a mixed-integer programming (MIP) approach to the optimal personalization tree problem (3).

We consider a fixed binary tree structure on nodes $1, \ldots, P$. Let $\mathcal{A}_p \subset [P]$ be the unique path from the root to node $p$, *i.e.*, its ancestors. For $q \in A_p$, let $\mathcal{R}_{pq} = 1$ if the right branch is taken to reach $p$ from $q$, otherwise $-1$. Let $\mathcal{L} = \{p \in [P] : p \notin \mathcal{A}(q) \,\forall q \in [P]\}$ be the set of

---

[2] Note that $I_{\text{pers}}(\mathcal{X})$ is only defined when there is at least 1 subject for each treatment in the partition $\mathcal{X}$. An alternative appropriate for scarce data and large $m$ allows for any number of subjects but chooses the best treatment only from among those with at least $n_{\text{min-leaf}}$ subjects in the partition.

---

**Algorithm 1** PT subroutine

**input:** Data part $\tilde{S} = \{(X_{i_1}, T_{i_1}, Y_{i_1}), \ldots, (X_{i_k}, T_{i_k}, Y_{i_k})\}$, current depth $\Delta$, tuning parameters $n_{\text{min-leaf}}, \Delta_{\max}, \#_{\text{features}}$.

1: **for** $\ell \in [d]$ **do** sort the data along $x_\ell$: $X_{i_{\pi(\ell,1)},\ell} \leq \cdots \leq X_{i_{\pi(\ell,k)},\ell}$.
2: Set $\hat{\tau}_{\tilde{S}}(x) = \arg\min_{t \in [m]} \sum_{j=1}^{k} \mathbb{I}[T_{i_j} = t] Y_{i_j} / \sum_{j=1}^{k} \mathbb{I}[T_{i_j} = t]$.
3: **if** $\Delta < \Delta_{\max}$ and $\min_{t \in [m]} \sum_{j=1}^{k} \mathbb{I}[T_{i_j} = t] > n_{\text{min-leaf}}$ **then**
4:  Set $I^\star = \infty, \ell^\star = 0, j^\star = 0$.
5:  Draw $\ell_1, \ldots, \ell_{\#_{\text{features}}}$ at random from $[d]$ without replacement.
6:  **for** $\ell = \ell_1, \ldots, \ell_{\#_{\text{features}}}$ **do**
7:   Set $k_1^{\text{L}} = \cdots = k_m^{\text{L}} = 0, S_1^{\text{L}} = \cdots = S_m^{\text{L}} = 0, k^{\text{L}} = 0$.
8:   Set $k_t^{\text{R}} = \sum_{j=1}^{k} \mathbb{I}[T_{i_j} = t], S_t^{\text{R}} = \sum_{j=1}^{k} \mathbb{I}[T_{i_j} = t] Y_{i_j}, k^{\text{R}} = k$.
9:   **for** $j \in [k-1]$ **do**
10:    Update $k^{\text{L}} += 1, k^{\text{R}} -= 1, t = T_{i_{\pi(\ell,j)}}, k_t^{\text{L}} += 1, k_t^{\text{R}} -= 1,$
11:    $S_t^{\text{L}} += Y_{i_{\pi(\ell,j)}}, S_t^{\text{R}} -= Y_{i_{\pi(\ell,j)}}$.
12:    Set $I = k^{\text{L}} \min_{t \in [m]} S_t^{\text{L}} / k_t^{\text{L}} + k^{\text{R}} \min_{t \in [m]} S_t^{\text{R}} / k_t^{\text{R}}$,
13:    $k_{\min} = \min_{t \in [m]} \min(k_t^{\text{L}}, k_t^{\text{R}})$.
14:    **if** $I < I^\star$ and $k_{\min} \geq n_{\text{min-leaf}}$ **then** set $I^\star = I, \ell^\star = \ell, j^\star = j$.
15:   **end for**
16:  **end for**
17:  **if** $I^\star < \infty$ **then**
18:   Set $\tilde{S}^{\text{L}} = \{(X_{i_{\pi(\ell^\star,j)}}, T_{i_{\pi(\ell^\star,j)}}, Y_{i_{\pi(\ell^\star,j)}}) : 1 \leq j \leq j^\star\}$,
19:   $\tilde{S}^{\text{R}} = \{(X_{i_{\pi(\ell^\star,j)}}, T_{i_{\pi(\ell^\star,j)}}, Y_{i_{\pi(\ell^\star,j)}}) : j^\star + 1 \leq j \leq k\}$,
20:   $\hat{\tau}_{\tilde{S}^{\text{L}}} = \text{Alg. 1}(\tilde{S}^{\text{L}}, \Delta + 1), \hat{\tau}_{\tilde{S}^{\text{R}}} = \text{Alg. 1}(\tilde{S}^{\text{R}}, \Delta + 1)$,
21:   $\theta^\star = \frac{1}{2}(X_{i_{\pi(\ell^\star,j)},\ell} + X_{i_{\pi(\ell^\star,j+1)},\ell})$,
22:   $\hat{\tau}_{\tilde{S}}(x) = (\textbf{if } x_{\ell^\star} \leq \theta^\star \textbf{ then } \hat{\tau}_{\tilde{S}^{\text{L}}}(x) \textbf{ else } \hat{\tau}_{\tilde{S}^{\text{R}}}(x))$.
23:  **end if**
24: **end if**

**output:** $\hat{\tau}_{\tilde{S}}(x)$.

---

leaf nodes and let $\mathcal{L}^C = [P] \setminus \mathcal{L}$ be the non-leaf nodes. Let $\mathcal{C}_p \subset [d] \times \mathbb{R}$ be the finite set of potential cuts at each non-leaf node $p \in \mathcal{L}^C$, where $(\ell, \theta) \in \mathcal{C}_p$ denotes that the cut $x_\ell \leq \theta$ is to be considered at node $p$. (Usually we take $\theta$ to be the data midpoints along dimension $\ell$.) Let $\overline{Y}_i = Y_i - \min_{j \in [n]} Y_j$, $\overline{Y}_{\max} = \max_i \overline{Y}_i$, and $M = \overline{Y}_{\max}(\max_{t \in [m]} \sum_{i=1}^{n} \mathbb{I}[T_i = t] - |\mathcal{L}| n_{\text{min-leaf}})$. For a vector $\gamma$ with index set $\mathcal{C} \subset [d] \times \mathbb{R}$, let $\chi_i(\gamma, \mathcal{C}) = \sum_{(\ell,\theta) \in \mathcal{C}} \mathbb{I}[X_{i,\ell} \leq \theta] \gamma_{\ell,\theta}$. For $p \in \mathcal{L}^C$, let $k_p = \lceil \log_2 |\mathcal{C}_p| \rceil$ and $Z_p \in \{0,1\}^{k_p \times |\mathcal{C}_p|}$ be such that $(Z_p)_{ij} = 1$ if $\lfloor j/2^i \rfloor$ is odd and otherwise 0. Our MIP formulation of the OPT problem (3) follows:

**minimize** $\sum_{i=1}^{n} \sum_{p \in \mathcal{L}} \nu_{ip}$ subject to (4a)

$w \in [0,1]^{[n] \times \mathcal{L}}, \lambda \in \{0,1\}^{\mathcal{L} \times m}, \mu \in \mathbb{R}_+^{\mathcal{L}}, \nu \in \mathbb{R}_+^{[n] \times \mathcal{L}}$ (4b)

$\gamma_p \in [0,1]^{\mathcal{C}_p}, \delta_p \in \{0,1\}^{k_p}, Z_p \gamma_p = \delta_p \,\forall p \in \mathcal{L}^C$ (4c)

$w_{ip} \leq \frac{1+\mathcal{R}_{pq}}{2} - \mathcal{R}_{pq} \chi_i(\gamma_q, \mathcal{C}_q) \,\forall i \in [n], p \in \mathcal{L}, q \in \mathcal{A}_p$ (4d)

$w_{ip} \geq 1 + \sum_{q \in \mathcal{A}_p} \left(\frac{1-\mathcal{R}_{pq}}{2} - \mathcal{R}_{pq} \chi_i(\gamma_q, \mathcal{C}_q)\right)$
$\qquad\qquad\qquad\qquad\qquad\qquad \forall i \in [n], p \in \mathcal{L}$ (4e)

$\sum_{i : T_i = t} w_{ip} \geq n_{\text{min-leaf}} \qquad \forall t \in [m]$ (4f)

$\nu_{ip} \leq \overline{Y}_{\max} w_{ip}, \nu_{ip} \leq \mu_p \qquad \forall i \in [n], p \in \mathcal{L}$ (4g)

$\nu_{ip} \geq \mu_p - \overline{Y}_{\max}(1 - w_{ip}) \qquad \forall i \in [n], p \in \mathcal{L}$ (4h)

$\sum_{t \in [m]} \lambda_{pt} = 1 \qquad \forall p \in \mathcal{L}$ (4i)

$\sum_{i : T_i = t} (\nu_{ip} - w_{ip} \overline{Y}_i) \leq M(1 - \lambda_{pt}) \quad \forall p \in \mathcal{L}, t \in [m]$ (4j)

$\sum_{i : T_i = t} (\nu_{ip} - w_{ip} \overline{Y}_i) \geq M(\lambda_{pt} - 1) \quad \forall p \in \mathcal{L}, t \in [m]$ (4k)

Problem (4) is a MIP with $|\mathcal{L}| m + \sum_{p \in \mathcal{L}^C} \log_2 |\mathcal{C}_p|$ binary variables. The variables $\gamma_p$ encode choice of cut at each node $p$ and constraint (4c) ensures only one is cho-

**Algorithm 2** PF

> **input:** Data $S_n = \{(X_1, T_1, Y_1), \ldots, (X_n, T_n, Y_n)\}$, tuning parameters $T$, $n_{\text{min-leaf}}$, $\Delta_{\max}$, #features.
> 1: **for** $j \in [T]$ **do**
> 2:    Draw $S_n^{(j)} = \{(X_{i_1}, T_{i_1}, Y_{i_1}), \ldots, (X_{i_n}, T_{i_n}, Y_{i_n})\}$ at random from $S_n$ with replacement.
> 3:    Set $\hat{\tau}_n^{(j)} = $ Alg. 1$(S_n^{(j)}, 0, n_{\text{min-leaf}}, \Delta_{\max}, \text{\#features})$.
> 4: **end for**
> **output:** $\hat{\tau}_n(x) = \text{mode}\{\hat{\tau}_n^{(1)}(x), \ldots, \hat{\tau}_n^{(T)}(x)\}$.

**Algorithm 3** OPT (complete binary tree)

> **input:** Data $S_n = \{(X_1, T_1, Y_1), \ldots, (X_n, T_n, Y_n)\}$, tuning parameters $n_{\text{min-leaf}}$, $\Delta$, #features, #cuts.
> 1: Set $P = 2^\Delta$, $\mathcal{L}^C = [2^{\Delta-1}]$, $\mathcal{A}_p = \{\lfloor p/2^j \rfloor : j \in [\Delta]\}$,
> 2:    $\mathcal{R}_{pq} = (-1)^{1+\lfloor p/2^{\Delta - \lfloor \log_2(q) \rfloor} \rfloor}$.
> 3: **for** $\ell \in [d]$ **do** sort the data along $x_\ell$: $X_{i_{\pi(\ell,1)},\ell} \leq \cdots \leq X_{i_{\pi(\ell,k)},\ell}$.
> 4: **for** $p = 1, \ldots, 2^{\Delta-1}$ **do**
> 5:    Draw $\mathcal{F}_p \subset [d]$ with $|\mathcal{F}_p| = $ #features. Set $J = \{1, \lceil \frac{n-1}{\text{\#cuts}} \rceil, \ldots, n-1\}$.
> 6:    Set $C_p = \{(\ell, \frac{X_{i_{\pi(\ell,j)},\ell} + X_{i_{\pi(\ell,j+1),\ell}}}{2}) : \ell \in \mathcal{F}_p, j \in J\}$.
> 7: **end for**
> 8: Find $\gamma$, $\lambda$ that solve problem (4).
> **output:** Personalization model $\hat{\tau}_n(x)$ that proceeds as follows:
> Set $p = 1$. **for** $j \in [\Delta]$ **do** set $(\ell, \theta) = \inf\{c \in \mathcal{C}_p : \gamma_{p,c} = 1\}$, $p = 2p + \mathbb{I}[x_\ell > \theta]$. **return** $\inf\{t \in [m] : \lambda_{pt} = 1\}$.

sen (see Yıldız & Vielma, 2013). The variable $w_{ip}$ encodes membership of datapoint $i$ to leaf $p$ and constraints (4d-4e) enforce that $w_{ip}$ is the product of indicators of whether $X_i$ goes in the left or right branch of the ancestor nodes. Since these constraints are integral (Ahuja et al., 1993) we need not enfoce $w_{ip}$ be binary. Constraint (4f) ensures at least $n_{\text{min-leaf}}$ samples per leaf. The variable $\mu_p$ encodes the mean outcome of the prescribed treatment in leaf $p$ and the variable $\nu_{ip}$ encodes its product with $w_{ip}$, as ensured by constraints (4g-4h). The variable $\lambda_{pt}$ encodes the choice of treatment $t$ in leaf $p$ and constraint (4i) ensures only one is chosen. Constraints (4j-4k) ensure the consistency between the choice of treatment $\lambda_{pt}$ and the mean outcome $\mu_p$. We summarize the OPT algorithm for a complete binary tree in Alg. 3. We use Gurobi to solve MIP (4) and use PT as a heuristic warm start, randomly splitting leaf nodes at depth less than $\Delta$.

### 2.6. Adapting Existing Methods to Personalization Using Observational Data

As discussed earlier, methods that estimate CATE, notably CT (Athey & Imbens, 2016), can be used to choose between two treatments by comparing $\delta(x) = \mathbb{E}[Y \mid X = x, T = 2] - \mathbb{E}[Y \mid X = x, T = 1]$ to zero. However, such methods are directed at estimation rather than personalization and only address two treatments. To address the latter, we propose *one-vs-all* (1vA) and *one-vs-one* (1v1) strategies for personalization.

For 1vA, for each $t \in [m]$ we learn an estimate $\hat{\delta}_n^{tvA}(x)$ of $\delta^{tvA}(x) = \mathbb{E}[Y \mid X = x, T = t] - \mathbb{E}[Y \mid X = x, T \neq t]$ by applying a base algorithm (*e.g.*, CT) to the modified dataset $S_n^{tvA} = \{(X_i, 1 + \mathbb{I}[T_i = t], Y_i) : i \in [n]\}$; then we assign the treatment that does the best compared to the rest: $\hat{\tau}_n^{1vA}(x) \in \arg\min_{t \in [m]} \hat{\delta}^{tvA}(x)$.

For 1v1, for each $t \neq s$ we learn an estimate $\hat{\delta}_{n_t+n_s}^{tvs}(x)$ of $\delta^{tvs}(x) = \mathbb{E}[Y \mid X = x, T = t] - \mathbb{E}[Y \mid X = x, T = s]$ on the modified data subset $S_{n_t+n_s}^{tvs} = \{(X_i, 1 + \mathbb{I}[T_i = t], Y_i) : T_i \in \{t, s\}\}$; then we either assign the treatment that does the best compared to the worst, $\hat{\tau}_n^{1v1\text{-}A}(x) \in \arg\min_{t \in [m]} \min_{s \in [m]} \hat{\delta}_{n_t+n_s}^{tvs}(x)$, or the one that gets the most votes in one-to-one comparisons, $\hat{\tau}_n^{1v1\text{-}B}(x) \in \arg\max_{t \in [m]} \sum_{s \neq t} \mathbb{I}\left[\hat{\delta}_{n_t+n_s}^{tvs}(x) < 0\right]$.

We can prove that each of our 1vA and 1v1 proposals are consistent given pointwise consistent estimates of CATE:

**Theorem 4.** *Let Asn. 1 hold. Then:*

1. *If $\hat{\delta}_n^{tvA}(X) \to \delta^{tvA}(X)$ a.s. $\forall t \in [m]$, then $\hat{\tau}_n^{1vA}(X) \in \mathcal{T}^*(X)$ eventually a.s.*
2. *If $\hat{\delta}_{n_t+n_s}^{tvs}(X) \to \delta^{tvs}(X)$ a.s. $\forall t \neq s$, then $\hat{\tau}_n^{1v1\text{-}A}(X), \hat{\tau}_n^{1v1\text{-}B}(X) \in \mathcal{T}^*(X)$ eventually a.s.*

Note that 1vA and 1v1 with CT do not inherit trees' interpretability because the partitions of the 1vX models may not overlap.

POEM and NPOEM (Swaminathan & Joachims, 2015a;b) solve the problem of learning from logged bandit feedback, assuming access to the logging policy that generated the data. To adapt these to personalizing from observational data, we propose to *impute* the logging policy using estimated GPS, *i.e.*, pretend the data were generated by the policy that assigns $t$ when context is $x$ with probability $\hat{\phi}_n(t, x)$ where $\hat{\phi}_n$ is a probabilistic classification model fitted to the data $\{(X_i, T_i) : i \in [n]\}$. We call these IPOEM and INPOEM.

## 3. Submatching for Validating Personalization using Observational Data

In this section we discuss how one can evaluate and validate personalization policies, such as the ones from the last section. Usually, a new policy would be evaluated using a randomized controlled trial, but these can be infeasibly costly. We consider how to evaluate a personalization policy using observational data. Such a dataset can be a subset removed from the training data either for the purpose of testing or for tuning and selection by (cross-)validation. The difficulty in using observational data is that if a policy prescribes any treatment $\tau(X_i) \neq T_i$, then it is not immediately clear how to score this.

For offline evaluation of contextual bandits with experimental data, Li et al. (2011) show that rejection sampling is sufficient. A similar solution to evaluation with obser-vational data is a combined rejection and importance sampling approach suggested by Thm. 2. If we had the GPS $Q_i$, we could omit any datapoint where $\tau(X_i) \neq T_i$ while giving score $Y_i/Q_i$ to each datapoint where the prescription matched the data, $\tau(X_i) = T_i$. Per Thm. 2 and the law of large numbers, this will provide a consistent estimate for out-of-sample personalization risk. However, not only does this throw away many datapoints, but to implement this in practice we would have to estimate the GPS from data. Estimating the GPS generally either relies heavily on model specification or, in non-parametric settings, can be biased and variable. This may be acceptable for training purposes, as in imputing GPS in IPOEM and INPOEM, as it is already a black box. However, for evaluation, a more reliable estimate of risk is desirable for evidence of success.

We propose the use of *submatching* for evaluation. Matching is a common tool for causal inference (Rosenbaum, 1989; Abadie & Imbens, 2006) where every subject is matched to a subject that received the opposite treatment, creating a complete matched dataset. In submatching, we instead seek only a *subset* of the data that is well-matched. In this subset, each subject is matched based on a metric $\|x - x'\|$ to $m - 1$ subjects that received each of the treatments the subject did not. Their outcome is imputed as the counterfactual outcome of those treatments for the subject. All matched subjects are not used for training in order to avoid in-sample bias. Usually, Mahalanobis distance is used: $((x-x')\hat\Sigma^{-1}(x-x'))^{1/2}$ where $\hat\Sigma$ is the sample covariance. (Note that due to personalization on $X$, matching on propensity scores alone would be insufficient.)

### 3.1. Greedy Submatching

The simplest way to extract a matched subset of size $n_{\text{test}}$ from $S_n$ is to do so *greedily*: draw random $i_1, \ldots, i_{n_{\text{test}}}$ from $[n]$ without replacement, for each $j \in [n_{\text{test}}]$ and $t \in [m]$, if $t = T_{i_j}$ then set $\hat Y_{i_j t} = Y_{i_j}$ and if $t \neq T_{i_j}$ then find $i \in \arg\min_{i \in [n]: T_i = t} \|X_i - X_{i_j}\|$ (with replacement), let $\hat Y_{i_j t} = Y_i$ and flag subject $i$, and finally remove all drawn and flagged subjects from training data. The imputed value for the *unknown* $Y_{i_j}(t)$ is $\hat Y_{i_j t}$ and our estimate for personalization risk of $\tau(x)$ is $\hat R(\tau) = \frac{1}{n_{\text{test}}}\sum_{j=1}^{n_{\text{test}}} \hat Y_{i_j \tau(X_{i_j})}$. When matching is exact, i.e. $X_i = X_{i_j}$ for all matches, this estimate is unbiased.

**Theorem 5.** *Under Asn. 1 and exact matching,* $\mathbb{E}[\hat R(\tau)] = R(\tau)$.

### 3.2. Optimal Submatching

The greedy method for constructing a matched dataset is simple but it can be wasteful, limiting the amount of the data available for training. We may be able to do better for testing and evaluation when $m = 2$, when the problem reduces to average treatment effect estimation. Consider the problem of finding the subset of the data with the closest matches. That is, finding $i_{11}, i_{12}, \ldots, i_{n_{\text{pair}}1}, i_{n_{\text{pair}}2}$ with $T_{i_{jt}} = t$ and minimal $\sum_{j=1}^{n_{\text{pair}}} \|X_{i_{j1}} - X_{i_{j2}}\|$, and using the pairs for imputations.[3] This problem can be reduced to bipartite matching, which can be solved efficiently (Hopcroft & Karp, 1973). Consider the complete bipartite graph with left nodes being subjects with $T_i = 1$ and right nodes being subjects with $T_i = 2$ along with $n - n_{\text{pair}}$ dummy nodes. Put weight $\|X_i - X_j\|$ on edges between datapoints and weight 0 on edges to dummy nodes. The subset of the data with the closest matches is given by the nodes incident to edges not incident to dummy nodes in the least-weight bipartite match. We extract these to construct a well-matched, economical test set with $n_{\text{test}} = 2n_{\text{pair}}$. Although this test set may be biased relative to the whole population (*e.g.*, it may emphasize areas of treatment overlap), the corresponding risk estimate is unbiased conditioned on the test set, *i.e.*, it corresponds to risk on an alternative population, which is often sufficient for comparison and selection.

### 3.3. Coefficient of Personalization

In prediction, the coefficient of determination $R^2$ is a unitless quantity bounded by 1 that measures both how well data $X$ predict outcomes $Y$ and how well a predictive model leverages $X$. One way to interpret out-of-sample $R^2$ is as the percent of the way that $X$ and the model go from a no-$X$-data solution ($Y$'s sample average) to perfect foresight ($Y$'s realized value). Using this interpretation, we construct two analogous quantities for personalization, the 1st and 2nd coefficients of personalization:

$$P_1(\tau) = 1 - \frac{\mathbb{E}[Y(\tau(X))] - \mathbb{E}[\min_{t\in[m]} Y(t)]}{\min_{t\in[m]} \mathbb{E}[Y(t)] - \mathbb{E}[\min_{t\in[m]} Y(t)]},$$
$$P_2(\tau) = 1 - \frac{\mathbb{E}[Y(\tau(X))] - \mathbb{E}[\min_{t\in[m]} Y(t)]}{\mathbb{E}[Y(T)] - \mathbb{E}[\min_{t\in[m]} Y(t)]}.$$

These are also analogous to the coefficient of prescription for conditional stochastic optimization (Bertsimas & Kallus, 2015). The first measures the improvement toward perfect (prescient) personalization relative to no personalization at all and the second does relative to current practice or standard of care (whatever determined $T$ in the data). They are unitless, bounded by 1. Using a matched dataset, we can estimate these as:[4]

$$\hat P_1(\tau) = 1 - \frac{\sum_{j=1}^{n_{\text{test}}} \hat Y_{i_j \tau(X_{i_j})} - \sum_{j=1}^{n_{\text{test}}} \min_{t\in[m]} \hat Y_{i_j t}}{\min_{t\in[m]} \sum_{j=1}^{n_{\text{test}}} \hat Y_{i_j t} - \sum_{j=1}^{n_{\text{test}}} \min_{t\in[m]} \hat Y_{i_j t}},$$
$$\hat P_2(\tau) = 1 - \frac{\sum_{j=1}^{n_{\text{test}}} \hat Y_{i_j \tau(X_{i_j})} - \sum_{j=1}^{n_{\text{test}}} \min_{t\in[m]} \hat Y_{i_j t}}{\sum_{j=1}^{n_{\text{test}}} Y_{i_j} - \sum_{j=1}^{n_{\text{test}}} \min_{t\in[m]} \hat Y_{i_j t}}.$$

---

[3] More efficient estimates may be possible using analogues of Robins & Rotnitzky (1995); Hahn (1998) on the submatched data.

[4] This assumes that potential outcomes are conditionally independent given $X$. Indeed, the conditional copula of potential outcomes has no physical meaning and is unidentifiable.

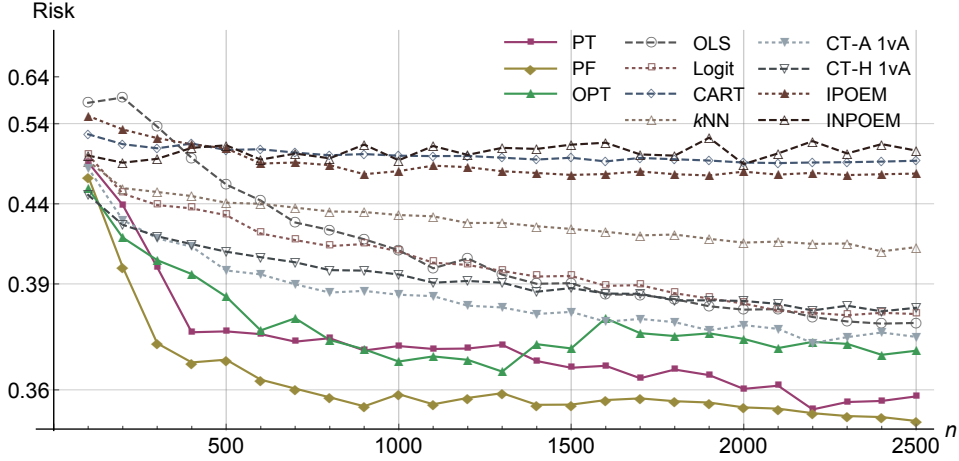

Figure 2. Personalization Risk for Personalized Warfarin Dosing

## 4. Empirical Investigation

We conclude with an empirical investigation of personalization using observational data and our new algorithms.

### 4.1. Personalized Warfarin Dosing

According to the International Warfarin Pharmacogenetics Consortium, "warfarin is the most widely used oral anticoagulant agent worldwide" and finding the appropriate dose is both difficult and important "because it can vary by a factor of 10 among patients" and "incorrect doses contribute to a high rate of adverse effects" (Consortium, 2009). Currently, the common practice is to start a new patient at 35 mg/week and slowly adjust the dose (Jaffer & Bragg, 2003). We present an application of our methods to personalizing dosage based on data on 5410 warfarin patients collected by Consortium (2009).

The baseline data collected on each patient include demographic characteristics (sex, ethnicity, age, weight, height, and smoker), reason for treatment (*e.g.*, atrial fibrillation), current medications, co-morbidities (*e.g.*, diabetes), genotype of two polymorphisms in CYP2C9, and genotype of seven single nucleotide polymorphisms (SNPs) in VKORC1. The correct stable therapeutic dose of warfarin, determined by adjustment over a few weeks, is recorded for each patient and segmented into three dose groups: low ($\leq 21$ mg/week, $t = 1$), medium ($> 21, < 49$ mg/week, $t = 2$), and high ($\geq 49$ mg/week, $t = 3$). The dataset was also studied in an online (bandit) setting in (Bastani & Bayati, 2016) where an R&C approach is analyzed.

In our experiment, we let $Y(t)$ be 1 if the dose $t$ is incorrect and otherwise 0. To generate observational data (where dosage is not revealed by experimentation), we consider $T$ chosen based on body mass index (BMI):

$$\mathbb{P}(T = t \mid X = x) = \frac{e^{(t-2)(x_{\text{BMI}} - \mu_{\text{BMI}})/\sigma_{\text{BMI}}}}{e^{-(x_{\text{BMI}} - \mu_{\text{BMI}})/\sigma_{\text{BMI}}} + 1 + e^{(x_{\text{BMI}} - \mu_{\text{BMI}})/\sigma_{\text{BMI}}}},$$

where $\mu_{\text{BMI}}$ and $\sigma_{\text{BMI}}$ are the sample mean and standard deviation of BMI. As an example, we run the PT algorithm with $\Delta_{\max} = 5$ on the whole data, generating the tree shown in Fig. 1. (We use $\Delta_{\max} = 5$ due to length constraints.) It is known that VKORC1 and CYP2C9 genotypes are strongly associated to warfarin dosage requirements (Li et al., 2006). PT is able to learn this relationship and it provides an efficient and interpretable dosing guideline where the effect of these genotypes is clear.

To assess the efficiency of various personalization algorithms, for each $n = 100, 200, \ldots, 2500$, we consider 100 replications in which we randomly select $n$ training subjects and $n_{\text{test}} = 2500$ test subjects (disjoint, without replacement). In each replication, we run 12 personalization algorithms and evaluate their risk on the test set (where full counterfactuals are available). We test standard R&C using four predictive models: OLS, logistic regression, CART (scikit-learn defaults), and $k$NN ($k = \lfloor \sqrt{n} \rfloor$). We compare these to our three direct personalization methods: PT ($n_{\text{min-leaf}} = 20, \Delta_{\max} = \infty, \#_{\text{features}} = d$), PF ($T = 500, n_{\text{min-leaf}} = 10, \Delta_{\max} = \infty, \#_{\text{features}} = \sqrt{d}$), and OPT ($n_{\text{min-leaf}} = 20, \#_{\text{features}} = d, \#_{\text{cuts}} = 10, \Delta = 2 + \mathbb{I}[n \geq 300]$, MIP solve time limited to 1 hour). We also compare to our 1vA strategy using Athey & Imbens (2016)'s CT-A (adaptive) and CT-H (honest with 50-50 split) and to IPOEM and INPOEM (parameters tuned on 25% holdout validation as in Swaminathan & Joachims, 2015a;b) with GPS imputed by cross-validated $\ell_1$-regularized multinomial regression using *R* package `glmnet`. (Due to limited space, we focus on 1vA, which outperformed 1v1.)

We plot the average risk over replicates in Fig. 2 (note log scale). It is evident that R&C approaches make inefficient use of the available data by splitting it and learning more than is necessary. While eventually reaching low risk ($< 0.4$), R&C using OLS and logistic regression take much longer ($n = 1300$) to get there than our direct methods, which achieve low risk very quickly ($n = 200$) and near-optimal risk ($\leq 0.36$) soon after ($n = 700$). Nonparametric R&C (CART, $k$NN), IPOEM, and INPOEM converge slowly. 1vA with CT-A and CT-H offers competitive performance for moderate $n$, but fails to achieve near-optimal risk even at $n = 2500$. CT-A offers a small edge over CT-H, which can be attributed to CT-H's splitting of the training data – indeed, CT-H's primary advantage are correctly sized confidence intervals, which are not used.

Among our direct methods, PF appears to work the best overall, for both small and large $n$, while PT achieve similar performance for $n \geq 2000$. For smaller $n$, OPT outperforms PT (and PF for $n = 100$) attributed to OPT's ability to find the best simple tree to fit the scarce data. For larger $n$, the MIP becomes so large that Gurobi is hardly able to improve the PT warm start, which has very limited depth. Therefore, we see performance deteriorate. We conclude OPT is best either for small datasets or for finding models that are reasonably efficient while being *exceedingly* simple and interpretable (depth 2–3 compared to depth 9–13 for PT at $n = 2500$), albeit at computational cost. Our best out-of-sample risk is 0.356, which translates to $\hat{P}_1 = 0.22$, $\hat{P}_2 = 0.47$: *i.e.*, 22% (or, 47%) of the way from no personalization (or, standard of care) to *perfect* personalization.

### 4.2. Personalized Job Training

We consider an application to personalized recommendations for a job training program. We use data from the National Supported Work Demonstration (LaLonde, 1986) (combining the experimental sample of 465 subjects with the 2490 PSID controls to create an observational dataset). The data includes 2935 individuals, 185 of which received a job training program in 1976-77 ($T_i = 1$). The data includes information about age, education level, ethnicity, marital status, earnings in years 1974-75, and earnings in 1978. This data is the standard benchmark in evaluation of causal methodologies for estimating an average treatment effect (Dehejia & Wahba, 2002). We consider an alternative setting where we give a personalized recommendation as to whether to enroll in the job training program, assuming enrollment costs $2,000. Therefore, we let $Y_i$ equal 1978 earnings less $2,000 if $T_i = 1$.

From the 2935, we extract an *optimal* matched test set of 55 pairs ($n_{\text{test}} = 110$) perfectly matched in all covariates except for a mean absolute deviation of $12 and $17 in 1974 and 1975 earnings, respectively, within pairs. On the remaining $n = 2825$ subjects, we train the same personalization models as above with the following changes: we omit logistic regression (outcomes not binary), use $n_{\text{min-leaf}} = 10$ for PT and OPT and $= 1$ for PF, use $\Delta = 4$ for OPT and let the MIP solve for 24 hours, use logistic regressions to im-pute GPS for IPOEM and INPOEM, and include the causal forest (CF) extension (Wager & Athey, 2017) of CT as implemented by the *R* package `gradient.forest`.

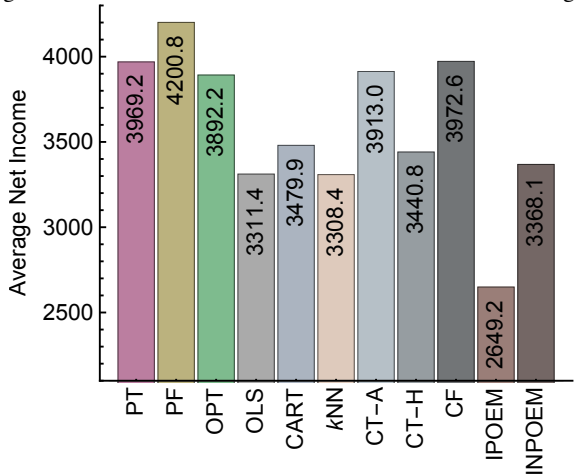

*Figure 3.* Personalization Benefit for Personalized Job Training

We plot the estimated average personalized net income (after enrollment costs) in Fig. 3. We see a clear benefit to our methods' direct targeting of personalization and that, with only two treatments, CF and CT-A provide highly competitive performance. Average net income of 4200.8 due to PF translates to $\hat{P}_1 = 0.22$, $\hat{P}_2 = 0.28$: *i.e.*, 22% (or, 28%) of the way from no personalization (or, standard of care) to *perfect* personalization.

## 5. Conclusions

We developed a new approach to the unique problem of personalization from observational data. The approach was based on a new formulation of the problem and a new impurity measure for personalization. This lead to three recursive-partitioning-based algorithms: the personalization tree that greedily partitions the data to minimize the sum of within-partition personalization impurities, the personalization forest that bagged many personalization trees, and the optimal personalization tree that used a MIP to globally optimize the partitioning problem. We developed new submatching techniques to evaluate and validate these algorithms as well as ones adapted from existing methods. And we used these techniques to evaluate all algorithms in two example applications: personalized warfarin dosing and personalized job training. In both examples, we demonstrated the benefits of our algorithms in terms of efficacy and interpretability. We phrased the success of our personalization techniques in terms of the new coefficients of personalization, which quantify the benefit we achieve from personalization as a percentage of the benefit that impossibly perfect personalization can achieve relative to either no personalization or the standard of care.

# Supplementary Material for: Recursive Partitioning for Personalization using Observational Data

Nathan Kallus [1]

## Omitted Proofs

*Proof of Theorem 1.* By Asn. 1, we have

$$\begin{aligned}\mathbb{E}\left[Y \mid X=x, T=t\right] &= \mathbb{E}\left[Y(T) \mid X=x, T=t\right] &&\text{(definition of } Y=Y(T)\text{)}\\ &= \mathbb{E}\left[Y(t) \mid X=x, T=t\right] &&\text{(conditioned on } T=t\text{)}\\ &= \mathbb{E}\left[Y(t) \mid X=x\right] &&\text{(Asn. 1).}\end{aligned}$$

Consider a realization of the data and $X = x$ where convergence occurs for all $t \in [m]$. Let

$$\epsilon(x) = \inf\{\zeta : s \in [m], \zeta = \left(\mathbb{E}\left[Y \mid X=x, T=s\right] - \min_{t \in [m]} \mathbb{E}\left[Y \mid X=x, T=t\right]\right) > 0\},$$

where $\inf(\varnothing) = \infty$. By assumption of convergence at this realization of the data and $X = x$, we have that eventually for all $t \in [m]$, $|\hat{\mu}_{t,n_t}(x) - \mathbb{E}\left[Y \mid X=x, T=t\right]| < \epsilon(x)/2$, at which point we must necessarily also have $\hat{\tau}_n(x) \in \arg\min_{t \in [m]} \mathbb{E}\left[Y \mid X=x, T=t\right] = \arg\min_{t \in [m]} \mathbb{E}\left[Y(t) \mid X=x\right]$. By assumption of pointwise consistency and because the intersection of finitely many a.s. events is a.s., the set of such realization of the data and $X = x$ have probability 1. $\square$

*Proof of Theorem 2.* First note that, given any $x$ with $\mathbb{P}\left(T=t \mid X=x\right) > 0$, we have

$$\mathbb{E}\left[Y \mid X=x, T=t\right] = \tfrac{\mathbb{E}[Y\mathbb{I}[T=t]|X=x]}{\mathbb{P}(T=t|X=x)} = \mathbb{E}\left[\tfrac{Y\mathbb{I}[T=t]}{\phi(t,x)} \mid X=x\right] = \mathbb{E}\left[\tfrac{Y\mathbb{I}[T=t]}{\phi(T,X)} \mid X=x\right] = \mathbb{E}\left[\tfrac{Y\mathbb{I}[T=t]}{Q} \mid X=x\right].$$

Therefore, since $\mathbb{P}\left(T=t \mid X\right) > 0$ almost surely,

$$\begin{aligned}R(\tau) = \mathbb{E}\left[Y(\tau(X))\right] &= \mathbb{E}\left[\mathbb{E}\left[Y(\tau(X)) \mid X\right]\right] &&\text{(iterated expectations)}\\ &= \mathbb{E}\left[\mathbb{E}\left[Y(\tau(X)) \mid X, T=\tau(X)\right]\right] &&\text{(Asn. 1)}\\ &= \mathbb{E}\left[\mathbb{E}\left[Y \mid X, T=\tau(X)\right]\right] &&\text{(definition of } Y\text{)}\\ &= \mathbb{E}\left[\mathbb{E}\left[Y\mathbb{I}[T=\tau(X)]/Q \mid X\right]\right] &&\text{(above observation)}\\ &= \mathbb{E}\left[Y\mathbb{I}[T=\tau(X)]/Q\right] &&\text{(iterated expectations)}. \quad\square\end{aligned}$$

*Proof of Theorem 4.* We start with 1vA. Restrict to $x$ such that $\phi(s,x) > 0\;\forall s$ (almost everywhere). Let $\mu(t,x) = \mathbb{E}\left[Y(t) \mid X=x\right]$. Under Asn. 1,

$$\begin{aligned}\delta^{tvA}(x) &= \mathbb{E}\left[Y \mid X=x, T=t\right] - \mathbb{E}\left[Y \mid X=x, T \neq t\right]\\ &= \mathbb{E}\left[Y \mid X=x, T=t\right] - \sum_{s \neq t} \mathbb{E}\left[Y \mid X=x, T=s\right]\mathbb{P}\left(T=s \mid X=x, T \neq t\right)\\ &= \mu(t,x) - \sum_{s \neq t} \phi(s,x)\mu(s,x)/\sum_{s \neq t}\phi(s,x).\end{aligned}$$

---

Since $\phi(s,x) > 0$, it's clear that $\delta^{tvA}(x) \leq \delta^{svA}(x)$ $\forall s$ if and only if $\mu(t,x) \leq \mu(s,x)$ $\forall s$. The rest of the proof for 1vA follows the same way as Thm. 1, showing that, under the assumption of pointwise consistent estimation, the estimation gap $\sup_{t \in [m]} \left| \hat{\delta}_n^{tvA}(x) - \delta^{tvA}(x) \right|$ is eventually smaller than half the decision gap, $\epsilon^{1vA}(x) = \inf\{\zeta : s \in [m], \zeta = \left(\delta^{svA}(x) - \min_{t \in [m]} \delta^{tvA}(x)\right) > 0\}$, a.s. and for almost everywhere $x$.

Next, we deal with 1v1-A. Fix $x$. Fix any $t_m \in \arg\max_{t \in [m]} \mu(t,x)$. Let $\delta^{tvmin}(x) = \min_{s \neq t} \delta^{tvs}(x)$. If $t, s \neq t_m$, then $\delta^{tvmin}(x) - \delta^{svmin}(x) = \mu(t,x) - \mu(s,x)$. On the other hand, for any $t \in [m]$, we always have both $\mu(t,x) - \mu(t_m, x) \leq 0$ and $\delta^{tvmin}(x) - \delta^{t_m vmin}(x) \leq 0$. Therefore, we have

$$\begin{aligned} t \in \arg\min_{t \in [m]} \mu(t,x) &\iff \mu(t,x) - \mu(s,x) \leq 0 \, \forall s \neq t \iff \mu(t,x) - \mu(s,x) \leq 0 \, \forall s \neq t, t_m \\ &\iff \delta^{tvmin}(x) - \delta^{svmin}(x) \leq 0 \, \forall s \neq t, t_m \iff \delta^{tvmin}(x) - \delta^{svmin}(x) \leq 0 \, \forall s \neq t \\ &\iff t \in \arg\min_{t \in [m]} \delta^{tvmin}(x). \end{aligned}$$

Let $\hat{\delta}_n^{tvmin}(x) = \min_{s \neq t} \hat{\delta}_{n_t + n_s}^{tvs}(x)$ and note that $\sup_{t \in [m]} \left| \hat{\delta}_n^{tvmin}(x) - \delta^{tvmin}(x) \right| \leq \sup_{t \in [m], s \in [m]} \left| \hat{\delta}_{n_t + n_s}^{tvs}(x) - \delta^{tvs}(x) \right|$, which converges to zero under pointwise consistency. The rest of the proof for 1v1-A follows as above, showing that this estimation gap is eventually smaller than half the decision gap, $\epsilon^{1v1-A}(x) = \inf\{\zeta : s \in [m], \zeta = \left(\delta^{svmin}(x) - \min_{t \in [m]} \delta^{tvmin}(x)\right) > 0\}$, a.s. and for almost everywhere $x$.

Next, we deal with 1v1-B. Fix $x$ and a realization of the data where convergence holds for all $t \neq s$. Then, eventually $\left| \hat{\delta}_{n_t + n_s}^{tvs}(x) - \delta^{tvs}(x) \right| \leq |\delta^{tvs}(x)|/2$ for all $t \neq s$ such that $\delta^{tvs}(x) \neq 0$. That is, eventually $\mathbb{I}\left[\hat{\delta}_{n_t + n_s}^{tvs}(x) < 0\right] = \mathbb{I}[\delta^{tvs}(x) < 0]$ for all $t \neq s$ such that $\delta^{tvs}(x) \neq 0$. Restrict to such large enough $n$. Let $k_t(x) = \sum_{t \neq s} \mathbb{I}[\delta^{tvs}(x) < 0]$, $\hat{k}_t(x) = \sum_{t \neq s} \mathbb{I}\left[\hat{\delta}_{n_t + n_s}^{tvs}(x) < 0\right]$, and $k_{\min}(x) = |\arg\min_{t \in [m]} \mu(t,x)|$. Then, $t \in \arg\min_{t \in [m]} \mu(t,x) \iff k_t(x) = m - k_{\min}(x) \iff \hat{k}_t(x) \geq m - k_{\min}(x) \Longleftarrow t \in \arg\max_{t \in [m]} \sum_{s \neq t} \mathbb{I}\left[\hat{\delta}_{n_t + n_s}^{tvs}(x) < 0\right]$. □

*Proof of Theorem 5.* By random sampling, $(X_{i_j}, T_{i_j}, Y_{i_j}(1), \ldots, Y_{i_j}(m))$ are distributed iid as $(X, T, Y(1), \ldots, Y(m))$ is in population. For $j \in [n_{\text{test}}]$, let $i_{jt}$ be $i_j$'s match for treatment $t$, or $i_j$ if $T_{i_j} = t$. Under exact matching, $Y_{i_{jt}}(1), \ldots, Y_{i_{jt}}(m) \mid X_{j_i}$ is distributed the same as $Y_{i_j}(1), \ldots, Y_{i_j}(m) \mid X_{j_i}, T_{j_i} = t$. By writing $\hat{Y}_{i_j t} = Y_{i_{jt}} = \sum_{s=1}^m \mathbb{I}[t = s] Y_{i_{js}}(s)$, we see that

$$\begin{aligned} \mathbb{E}[\hat{Y}_{i_j \tau(X_{i_j})}] &= \mathbb{E}\left[\mathbb{E}\left[\sum_{s=1}^m \mathbb{I}\left[s = \tau(X_{i_j})\right] Y_{i_{js}}(s) \mid X_{i_j}\right]\right] & \text{(iterated expectation)} \\ &= \sum_{s=1}^m \mathbb{E}\left[\mathbb{I}\left[s = \tau(X_{i_j})\right] \mathbb{E}\left[Y_{i_{js}}(s) \mid X_{j_i}\right]\right] & \text{(linearity)} \\ &= \sum_{s=1}^m \mathbb{E}\left[\mathbb{I}[s = \tau(X_i)] \mathbb{E}\left[Y_{i_j}(s) \mid X_i, T_i = s\right]\right] & \text{(exact matching)} \\ &= \sum_{s=1}^m \mathbb{E}\left[\mathbb{I}[s = \tau(X_i)] \mathbb{E}\left[Y_{i_j}(s) \mid X_i\right]\right] & \text{(Asn. 1)} \\ &= \mathbb{E}\left[\mathbb{E}\left[\sum_{s=1}^m \mathbb{I}[s = \tau(X_i)] Y_{i_j}(s) \mid X_i\right]\right] & \text{(linearity)} \\ &= \mathbb{E}\left[Y_{i_j}(\tau(X_{i_j}))\right] & \text{(iterated expectation)} \quad \square \end{aligned}$$